\documentclass[sigplan,screen]{acmart}
\renewcommand\footnotetextcopyrightpermission[1]{} 
\setcopyright{acmcopyright}
\acmPrice{15.00}
\acmDOI{10.1145/3314221.3314638}
\acmYear{2019}
\copyrightyear{2019}
\acmISBN{978-1-4503-6712-7/19/06}
\acmConference[PLDI '19]{Proceedings of the 40th ACM SIGPLAN Conference on Programming Language Design and Implementation}{June 22--26, 2019}{Phoenix, AZ, USA}
\acmBooktitle{Proceedings of the 40th ACM SIGPLAN Conference on Programming Language Design and Implementation (PLDI '19), June 22--26, 2019, Phoenix, AZ, USA}

\usepackage{booktabs}   
\usepackage{subcaption} 

\usepackage[linesnumbered,norelsize]{algorithm2e}
                        
\usepackage[utf8]{inputenc}

\usepackage{enumitem}

\usepackage{stmaryrd}

\usepackage{amsmath}

\usepackage{thmtools}

\usepackage{multirow}

\usepackage{flushend}

\usepackage{colortbl} 

\usepackage{xcolor} 

\usepackage{todonotes}

\DeclareFixedFont{\ttb}{T1}{txtt}{bx}{n}{12} 
\DeclareFixedFont{\ttm}{T1}{txtt}{m}{n}{12}  

\usepackage{color}
\definecolor{deepblue}{rgb}{0,0,0.5}
\definecolor{deepred}{rgb}{0.6,0,0}
\definecolor{deepgreen}{rgb}{0,0.5,0}

\usepackage{listings}

\newcommand\pythonstyle{\lstset{
language=Python,
basicstyle=\ttfamily\scriptsize,,
otherkeywords={self},             
keywordstyle=\color{deepblue}\ttfamily,
emph={MyClass,__init__},          
emphstyle=\ttb\color{deepred},    
stringstyle=\color{green}\ttfamily,
commentstyle=\color{red}\ttfamily,
frame=tb,                         
showstringspaces=false            %
}}

\lstnewenvironment{python}[1][]
{
\pythonstyle
\lstset{#1}
}
{}

\newcommand\pythoninline[1]{{\pythonstyle\lstinline!#1!}}

\newcommand{\cal}{\mathcal}
\DeclareMathOperator*{\argmaxA}{arg\,max}

\newcommand{\norm}[1]{\left\lVert#1\right\rVert}

\begin{document}

\title[An Inductive Synthesis Framework for Verifiable Reinforcement Learning]
{An Inductive Synthesis Framework for Verifiable {Reinforcement} Learning}         


\author{He Zhu}
\affiliation{
  \institution{Galois, Inc.}            
}
\author{Zikang Xiong}
\affiliation{
  \institution{Purdue University}            
}
\author{Stephen Magill}
\affiliation{
  \institution{Galois, Inc.}            
}
\author{Suresh Jagannathan}
\affiliation{
  \institution{Purdue University}            
}

\begin{abstract}
  Despite the tremendous advances that have been made in the last
  decade on developing useful machine-learning applications, their
  wider adoption has been hindered by the lack of strong assurance
  guarantees that can be made about their behavior.  In this paper, we
  consider how formal verification techniques developed for
  traditional software systems can be repurposed for verification of
  reinforcement learning-enabled ones, a particularly important class
  of machine learning systems.  Rather than enforcing safety by
  examining and altering the structure of a complex neural network
  implementation, our technique uses blackbox methods to synthesizes
  deterministic programs, simpler, more interpretable, approximations
  of the network that can nonetheless guarantee desired safety
  properties are preserved, even when the network is deployed in
  unanticipated or previously unobserved environments.  Our
  methodology frames the problem of neural network verification in
  terms of a counterexample and syntax-guided inductive synthesis
  procedure over these programs.  The synthesis procedure searches for
  both a deterministic program and an inductive invariant over an
  infinite state transition system that represents a specification of
  an application's control logic.  Additional specifications defining
  environment-based constraints can also be provided to further refine
  the search space.  Synthesized programs deployed in conjunction with
  a neural network implementation dynamically enforce safety
  conditions by monitoring and preventing potentially unsafe actions
  proposed by neural policies.  Experimental results over a wide range
  of cyber-physical applications demonstrate that
  software-inspired formal verification techniques can be used to
  realize trustworthy reinforcement learning systems with low overhead.
\end{abstract}

%


\maketitle

\section{Introduction}

Neural networks have proven to be a promising software architecture
for expressing a variety of machine learning applications.  However,
non-linearity and stochasticity inherent in their design greatly
complicate reasoning about their behavior.  Many existing approaches
to verifying~\cite{reluplex, AI2, AI3, reachability} and
testing~\cite{DeepTest,DeepConcTest,featureguidedtesting} these
systems typically attempt to tackle implementations head-on, reasoning
directly over the structure of activation functions, hidden layers,
weights, biases, and other kinds of low-level artifacts that are
far-removed from the specifications they are intended to satisfy.
Moreover, the notion of safety verification that is typically
considered in these efforts ignore effects induced by the actual
environment in which the network is deployed, significantly weakening
the utility of any safety claims that are actually
proven. Consequently, effective verification methodologies in this
important domain still remains an open problem.

To overcome these difficulties, we define a new verification toolchain
that reasons about correctness extensionally, using a syntax-guided
synthesis framework~\cite{sygus} that generates a simpler and more
malleable deterministic program guaranteed to represent a safe control
policy of a reinforcement learning (RL)-based neural network, an
important class of machine learning systems, commonly used to govern
cyber-physical systems such as autonomous vehicles, where high
assurance is particularly important.  {Our synthesis
procedure is designed with verification in mind, and is thus
structured to incorporate formal safety constraints drawn from a
logical specification of the control system the network purports to
implement, along with additional salient environment properties
relevant to the deployment context.}
Our synthesis procedure 
treats the neural network as an oracle, extracting a deterministic program 
${\cal P}$ intended to approximate the policy actions implemented by 
the network.
{Moreover, our procedure ensures that a synthesized 
program ${\cal P}$ is formally verified safe.}
To this end, we realize our synthesis procedure via a \emph{counterexample 
guided inductive synthesis} (CEGIS) loop~\cite{sygus} that eliminates any
counterexamples to safety of ${\cal P}$.  
{More importantly, rather than repairing the network directly to 
satisfy the constraints governing ${\cal P}$, we instead treat ${\cal P}$ as 
a safety shield that operates in tandem with the network, overriding 
network-proposed actions whenever such actions can be shown to lead to 
a potentially unsafe state.  Our shielding mechanism thus retains performance, 
provided by the neural policy, while maintaining safety, provided by the program.}
Our approach naturally generalizes to \emph{infinite state} systems 
with a \emph{continuous} underlying action space.
Taken together, these properties enable safety enforcement of RL-based
neural networks without having to suffer a loss in performance to
achieve high assurance.  We show that over a range of cyber-physical
applications defining various kinds of control systems, the overhead
of runtime assurance is nominal, less than a few percent, compared to
running an unproven, and thus potentially unsafe, network with no
shield support.
This paper makes the following contributions:
\begin{enumerate}
\item We present a verification toolchain for ensuring that the
  control policies learned by an RL-based neural network are safe.
  Our notion of safety is defined in terms of a specification of an
  infinite state transition system that captures, for example, the
  system dynamics of a cyber-physical controller.

\item We develop a counterexample-guided inductive synthesis framework 
  that treats the neural control policy as an oracle to guide the search 
  for a simpler deterministic program that approximates the behavior of 
  the network but which is more amenable for verification.  The synthesis
  procedure differs from prior efforts~\cite{pirl,qdagger} because the
  search procedure is bounded by safety constraints defined by the
  specification (i.e., state transition system) as well as a
  characterization of specific environment conditions defining the
  application's deployment context.

\item We use a verification procedure that guarantees actions proposed
  by the synthesized program always lead to a state consistent with an
  inductive invariant of the original specification and deployed
  environment context.  This invariant defines an inductive property
  that separates all reachable (safe) and unreachable (unsafe) states
  expressible in the transition system.

\item We develop a runtime monitoring framework that treats the
  synthesized program as a safety shield~\cite{AB+18}, overriding
  proposed actions of the network whenever such actions can cause the
  system to enter into an unsafe region.
\end{enumerate}

We present a detailed experimental study over a wide range of
  cyber-physical control systems that justify the utility of our
  approach.  These results indicate that the cost of ensuring
  verification is low, typically on the order of a few percent.
The remainder of the paper is structured as follows.  In the next
section, we present a detailed overview of our approach.
Sec.~\ref{sec:problem} formalizes the problem and the context.
Details about the synthesis and verification procedure are given in
Sec.~\ref{sec:verificationprocedure}. A detailed evaluation study is
provided in Sec.~\ref{sec:eval}.  Related work and conclusions are
given in Secs.~\ref{sec:related} and~\ref{sec:conc}, resp.

\section{Motivation and Overview}

To motivate the problem and to provide an overview of our approach,
consider the definition of a learning-enabled controller that operates
an inverted pendulum. While the specification of this system is
simple, it is nonetheless representative of a number of practical
control systems, such as Segway transporters and autonomous drones,
that have thus far proven difficult to verify, but for which high
assurance is very desirable.

\subsection{State Transition System}

\begin{figure}[t]
\centering
\includegraphics[width=0.37\textwidth]{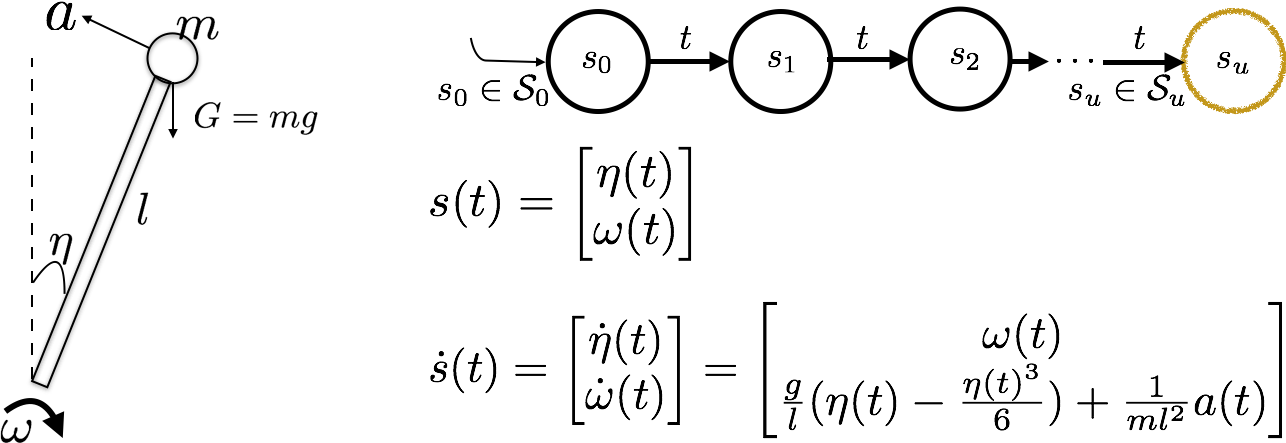}
\caption{Inverted Pendulum State Transition System.
The pendulum has mass $m$ and length $l$.
A system state is $s = [\eta, \omega]^T$ where $\eta$ is the 
its angle and $\omega$ is its angular
velocity.  A continuous control action $a$ 
maintains the pendulum upright.
} \label{fig:st}
\end{figure}

We model an inverted pendulum system as an \emph{infinite} 
state transition system with \emph{continuous} actions in Fig.~\ref{fig:st}.  
Here, the pendulum has mass $m$ and length $l$.
A system state is $s = [\eta, \omega]^T$ where $\eta$ is the 
pendulum's angle and $\omega$ is its angular
velocity.  A controller can use a 1-dimensional 
continuous control action $a$ to maintain the pendulum upright.

Since modern controllers are typically implemented digitally (using
digital-to-analog converters for interfacing between the analog system
and a digital controller), we assume that the pendulum is controlled
in discrete time instants $kt$ where $k=0,1,2,\cdots$, \emph{i.e.},
the controller uses the system dynamics, the change of rate of $s$,
denoted as $\dot{s}$, to transition every $t$ time period, with the
conjecture that the control action $a$ is a constant during each
discrete time interval.  Using Euler's method, for example, a
transition from state $s_k = s(kt)$ at time $kt$ to time $kt + t$ is
approximated as $s(kt+t) = s(kt) + \dot{s}(kt)\times t$.  We specify
the change of rate $\dot{s}$ using the differential equation shown in
Fig.~\ref{fig:st}.\footnote{We derive the control dynamics equations
  assuming that an inverted pendulum satisfies general Lagrangian
  mechanics and approximate non-polynomial expressions with their
  Taylor expansions.}  Intuitively, the control action $a$ is allowed
to affect the change of rate of $\eta$ and $\omega$ to balance a
pendulum.  Thus, small values of $a$ result in small swing and
velocity changes of the pendulum, actions that are useful when the
pendulum is upright (or nearly so), while large values of $a$
contribute to larger changes in swing and velocity, actions that are
necessary when the pendulum enters a state where it may risk losing
balance.  In case the system dynamics are unknown, we can use known
algorithms to infer dynamics from online
experiments~\cite{reza-clancy}.

Assume the state transition system of the inverted pendulum starts
from a set of initial states $\cal{S}_0$:
\[\cal{S}_0: \{(\eta, \omega)\ \vert -20^\circ \le \eta \le 20^\circ \wedge
-20^\circ \le \omega \le 20^\circ\}\]
The global safety property we wish to preserve is that the pendulum never falls down.
We define a set of unsafe states $\cal{S}_u$ of the transition system
(colored in yellow in Fig.~\ref{fig:st}):
\[\cal{S}_u: \{(\eta, \omega)\ \vert\ \neg (-90^\circ < \eta < 90^\circ \wedge
-90^\circ < \omega < 90^\circ)\}\]

\noindent We assume the existence of a neural network control policy
$\pi_{\mathbf{w}}: \mathbb{R}^2 \rightarrow \mathbb{R}$ that executes
actions over the pendulum, whose weight values of $\mathbf{w}$ are
learned from training episodes.  This policy is a state-dependent
function, mapping a $2$-dimensional state $s$ ($\eta$ and $\omega$) to
a control action $a$.  At each transition, the policy mitigates
uncertainty through feedback over state variables in $s$.

\noindent{\bf Environment Context}. An environment context
$\cal{C}[\cdot]$ defines the behavior of the application, where
$[\cdot]$ is left open to deploy a reasonable neural controller
$\pi_{\mathbf{w}}$.  The actions of the controller are dictated by
constraints imposed by the environment.  In its simplest form, the
environment is simply a state transition system.  In the pendulum
example, this would be the equations given in Fig.~\ref{fig:st},
parameterized by pendulum mass and length.  In general, however, the
environment may include additional constraints (\emph{e.g.}, a
constraining bounding box that restricts the motion of the pendulum
beyond the specification given by the transition system in Fig.~\ref{fig:st}).

\subsection{Synthesis, Verification and Shielding}
\label{sec:safetyverification}


In this paper, we assume that a neural network is trained using a 
state-of-the-art reinforcement learning strategy~\cite{dpg,ddpg}.  
Even though the resulting neural policy may appear to work well 
in practice, the complexity of its implementation makes it difficult to 
assert any strong and provable claims about its correctness 
{since the neurons, layers, weights and biases are 
far-removed from the intent of the actual controller.
We found that state-of-the-art neural network verifiers~\cite{reluplex,AI2}
are ineffective for verification of a neural controller 
over an infinite time horizon with complex system dynamics.
}


\begin{figure}[t]
\centering
\includegraphics[width=0.38\textwidth]{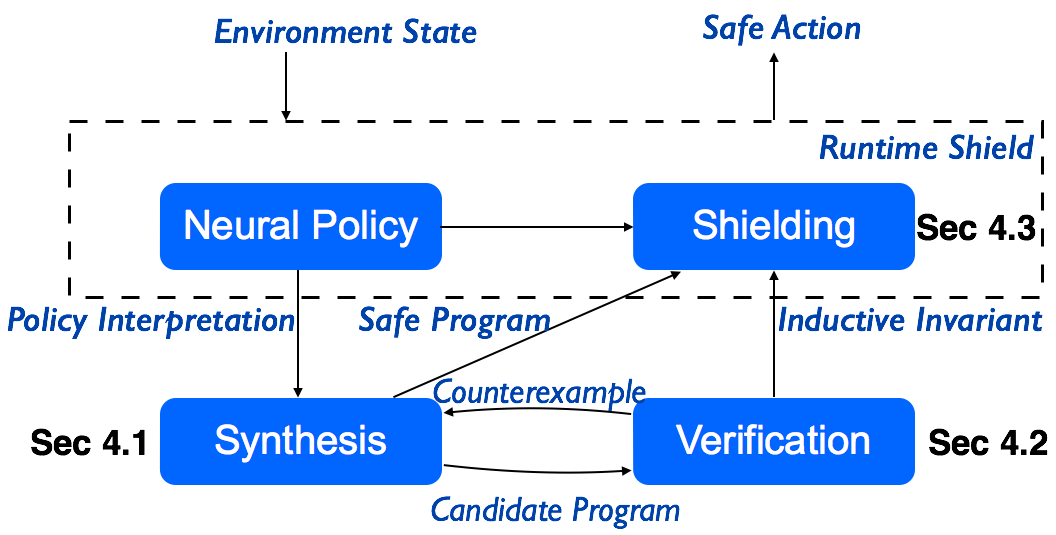}
\caption{The Framework of Our Approach.} \label{fig:framework}
\end{figure}

\noindent{\bf Framework.}  We construct a \emph{policy interpretation}
mechanism to enable verification, inspired by prior work on imitation
learning~\cite{imitationlearning,dagger} and interpretable machine
learning~\cite{pirl,qdagger}.
Fig.~\ref{fig:framework} depicts the high-level framework of our
approach.  Our idea is to synthesize a \emph{deterministic policy program}
from a neural policy $\pi_{\mathbf{w}}$, approximating $\pi_{\mathbf{w}}$ 
(which we call an \emph{oracle}) with a simpler structural program $\cal{P}$.
Like $\pi_{\mathbf{w}}$, $\cal{P}$ takes as input a system state 
and generates a control action $a$.
To this end, ${\cal P}$ is simulated in the environment used to train the 
neural policy $\pi_{\mathbf{w}}$, to collect feasible states.  Guided by
$\pi_{\mathbf{w}}$'s actions on such collected states, $\cal{P}$ is
further improved to resemble $\pi_{\mathbf{w}}$.

The goal of the synthesis procedure is to search for a deterministic
program $\cal{P}^*$ satisfying both (1) a quantitative specification
such that it bears reasonably close resemblance to its oracle so that
allowing it to serve as a potential substitute is a sensible notion,
and (2) a desired logical safety property such that when in operation
the unsafe states defined in the environment $\cal{C}$ cannot be
reached.  Formally,
\begin{equation}\label{eq:goal}
\cal{P}^* = \argmaxA_{\cal{P} \in \text{Safe}(\cal{C}, \llbracket \cal{H} \rrbracket)} d(\pi_{\mathbf{w}}, \cal{P}, \cal{C})
\end{equation}
where $d(\pi_{\mathbf{w}}, \cal{P}, \cal{C})$ measures proximity of $\cal{P}$
with its neural oracle in an environment $\cal{C}$;
$\llbracket \cal{H} \rrbracket$ defines a search space for $\cal{P}$
with prior knowledge on the shape of target deterministic programs;
and, $\text{Safe}(\cal{C}, \llbracket \cal{H} \rrbracket)$ restricts the 
solution space to a set of safe programs. 
A program $\cal{P}$ is safe if the safety of the transition system 
$\cal{C}[\cal{P}]$, the deployment of $\cal{P}$ in the environment 
$\cal{C}$, can be \emph{formally verified}. 

\label{interp-comp}
The novelty of our approach against prior work on neural policy
interpretation~\cite{pirl,qdagger} is thus two-fold:
\begin{enumerate}[leftmargin=*]
\item We bake in the concept of safety and formal safety verification
  into the synthesis of a deterministic program from a neural policy
  as depicted in Fig.~\ref{fig:framework}. If a candidate program is
  not safe, we rely on a counterexample-guided inductive synthesis
  loop to improve our synthesis outcome to enforce the safety conditions
  imposed by the environment.
\item We allow ${\cal P}$ to operate in tandem with the
  high-performing neural policy.  ${\cal P}$ can be viewed as
  capturing an \emph{inductive} invariant of the state transition
  system, which can be used as a shield to describe a boundary of safe
  states within which the neural policy is free to make optimal
  control decisions.  If the system is about to be driven out of this
  safety boundary, the synthesized program is used to take an action
  that is guaranteed to stay within the space subsumed by the
  invariant.  By allowing the synthesis procedure to treat the neural
  policy as an oracle, we constrain the search space of feasible
  programs to be those whose actions reside within a proximate
  neighborhood of actions undertaken by the neural policy.
\end{enumerate}

\noindent{\bf Synthesis.} Reducing a complex neural policy to a
simpler yet safe deterministic program is possible because we do not
require other properties from the oracle; specifically, we do not
require that the deterministic program precisely mirrors the
performance of the neural policy.  For example, experiments described
in~\cite{linearpolicy} show that while a linear-policy controlled
robot can effectively stand up, it is unable to learn an efficient
walking gait, unlike a sufficiently-trained neural policy.  However,
if we just care about the safety of the neural network, we posit that
a linear reduction can be sufficiently expressive to describe
necessary safety constraints.  Based on this hypothesis, for our
inverted pendulum example, we can explore a \emph{linear} program
space from which a deterministic program $\cal{P}_\theta$ can be drawn
expressed in terms of the following program sketch:
\begin{center}
\begin{python}[mathescape=true]
def $\cal{P}$[$\theta_1$, $\theta_2$]($\eta$, $\omega$): return $\theta_1 \eta + \theta_2 \omega$
\end{python}
\end{center}
Here, $\theta = [\theta_1, \theta_2]$ are unknown parameters that need
to be synthesized.  Intuitively, the program weights the importance of
$\eta$ and $\omega$ at a particular state to provide a feedback
control action to mitigate the deviation of the inverted pendulum from
$(\eta = 0^\circ, \omega = 0^\circ)$.

Our search-based synthesis sets $\theta$ to {\bf 0} initially.  It
runs the deterministic program $\cal{P}_\theta$ instead of the oracle
neural policy $\pi_{\mathbf{w}}$ within the environment $\cal{C}$
defined in Fig.~\ref{fig:st} (in this case, the state transition 
system represents the differential equation specification of 
the controller) to collect a batch of trajectories.
A run of the state transition system of $\cal{C}[\cal{P}_\theta]$
produces a finite trajectory $s_0, s_1, \cdots, s_T$. 
We find $\theta$ from the following optimization task 
that realizes~\eqref{eq:goal}:
\begin{equation}\label{eq:synthesis}
\max_{\theta \in \mathbb{R}^2} \mathbb{E} [\Sigma^T_{t=0}d(\pi_{\mathbf{w}}, \cal{P}_\theta, s_t)]
\end{equation}
where $d(\pi_{\mathbf{w}}, \cal{P}_\theta, s_t) \equiv \begin{cases}-(\cal{P}_\theta(s_t) - \pi_{\mathbf{w}}(s_t))^2 & s_t \notin \cal{S}_u
\\ -MAX & s_t \in \cal{S}_u \end{cases}$.
This equation aims to search for a program $\cal{P}_\theta$ at minimal 
distance from the neural oracle $\pi_{\mathbf{w}}$ along sampled 
trajectories, while simultaneously maximizing the likelihood that 
$\cal{P}_\theta$ is safe. 

Our synthesis procedure described in Sec.~\ref{sec:synthesize} is a
random search-based optimization algorithm~\cite{randomsearchbasic}.
We sample a new position of $\theta$ iteratively from its hypersphere
of a given small radius surrounding the current position of $\theta$
and move to the new position (\emph{w.r.t.} a learning rate) as
dictated by Equation (2). For the running example, our search
synthesizes:
\begin{center}
\begin{python}[mathescape=true]
def $\cal{P}$($\eta$, $\omega$): return $-12.05 \eta + -5.87 \omega$
\end{python}
\end{center}
The synthesized program can be used to intuitively interpret how the
neural oracle works. For example, if a pendulum with a positive angle
$\eta > 0$ leans towards the right ($\omega > 0$), the controller will
need to generate a large negative control action to force the pendulum
to move left.

\noindent{\bf Verification.}  Since our method synthesizes
a deterministic program $\cal{P}$, we can leverage off-the-shelf formal
verification algorithms to verify its safety with respect to the state
transition system $\cal{C}$ defined in Fig.~\ref{fig:st}.  To ensure
that $\cal{P}$ is safe, we must ascertain that it can never
transition to an unsafe state, \emph{i.e.}, a state that causes the pendulum
to fall.  When framed as a formal verification problem, answering
such a question is tantamount to discovering an inductive invariant
$\varphi$ that represents all safe states over the state transition
system:
\setlist{nolistsep}
\begin{enumerate}[noitemsep]
\item \emph{Safe:} $\varphi$ is disjoint with all unsafe states $\cal{S}_u$,
\item \emph{Init:} $\varphi$ includes all initial states $\cal{S}_0$,
\item \emph{Induction:} Any state in $\varphi$ transitions to another
  state in $\varphi$ and hence cannot reach an unsafe state.
\end{enumerate}

\noindent Inspired by template-based constraint solving approaches on
inductive invariant inference~\cite{template1,template2,barriercertificate,huikong}, 
the verification algorithm described in Sec.~\ref{sec:verify} uses a
constraint solver to look for an inductive invariant in the
form of a \emph{convex barrier certificate}~\cite{barriercertificate} 
$E(s) \le 0$ that maps all the states in the (safe) reachable set to 
non-positive reals and all the states in the unsafe set to positive reals.  
The basic idea is to identify a polynomial function 
$E: \mathbb{R}^n \rightarrow \mathbb{R}$ such that 
1) $E(s) > 0$ for any state $s \in \cal{S}_u$, 
2) $E(s) \le 0$ for any state $s \in \cal{S}_0$,
and 3) $E(s') - E(s) \le 0$ for any state $s$ that
transitions to $s'$ in the state transition system $\cal{C}[\cal{P}]$.  
The second and third condition collectively guarantee that $E(s) \le
0$ for any state $s$ in the reachable set, thus implying that an
unsafe state in $\cal{S}_u$ can never be reached.  

\begin{figure}[t]
\centering
\includegraphics[width=0.45\textwidth]{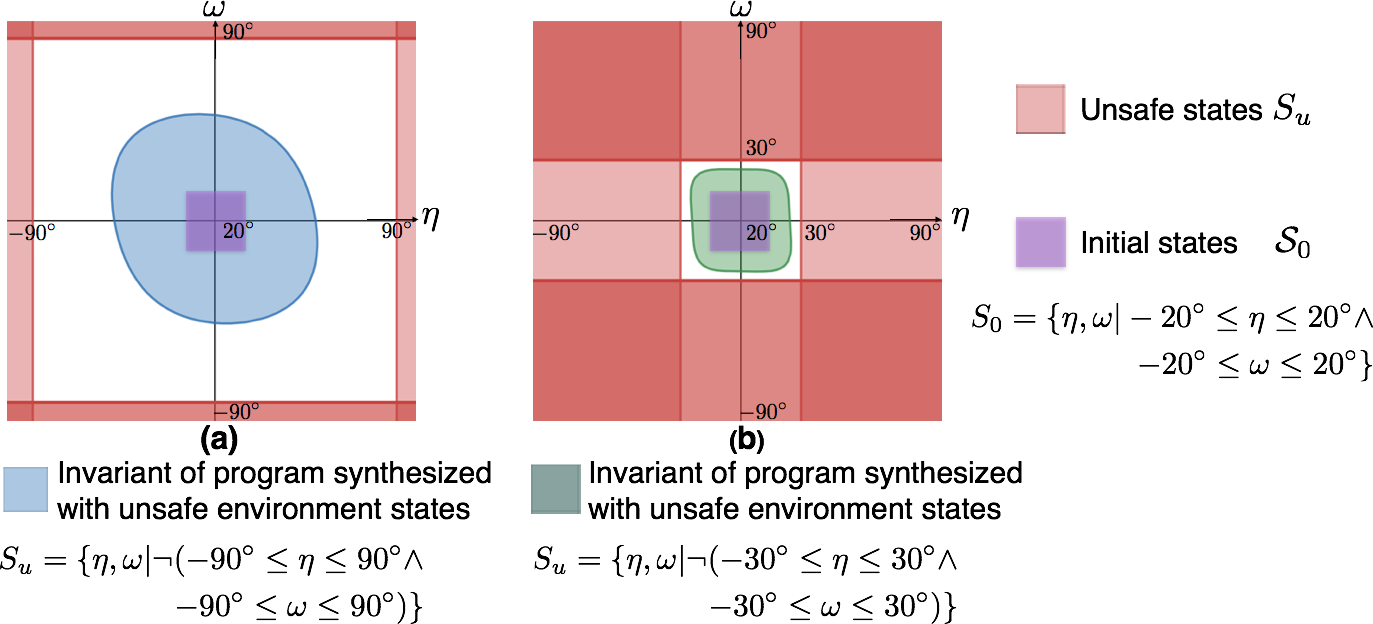}
\caption{Invariant Inference on Inverted Pendulum.} \label{fig:penduluminv}
\end{figure}

Fig.~\ref{fig:penduluminv}(a) draws the discovered invariant in blue
for $\cal{C}[\cal{P}]$ given the initial and unsafe states where
$\cal{P}$ is the synthesized program for the inverted
pendulum system.
We can conclude that the safety property is satisfied by the 
$\cal{P}$ controlled system as 
all reachable states do not overlap with unsafe states.
In case verification fails, our approach conducts a 
counterexample guided loop (Sec.~\ref{sec:verify}) to iteratively 
synthesize safe deterministic programs until verification succeeds. 

\noindent{\bf Shielding.}  Keep in mind that a safety proof of a
reduced deterministic program of a neural network does not
automatically lift to a safety argument of the neural network from
which it was derived since the network may exhibit behaviors not fully
captured by the simpler deterministic program.  To bridge this divide,
we propose to recover soundness at runtime by monitoring system
behaviors of a neural network in its actual environment (deployment)
context.

\begin{figure}[t]
\centering
\includegraphics[width=0.35\textwidth]{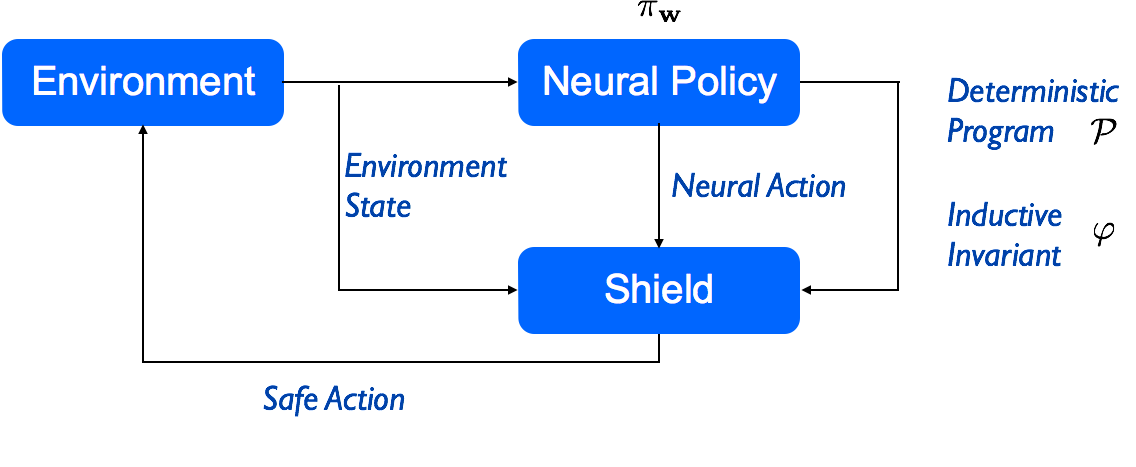}
\caption{The Framework of Neural Network Shielding.} \label{fig:neuralshield}
\end{figure}

Fig.~\ref{fig:neuralshield} depicts our runtime shielding approach
with more details given in Sec.~\ref{sec:shield}.  The inductive
invariant $\varphi$ learnt for a deterministic program $\cal{P}$ of a
neural network $\pi_{\mathbf{w}}$ can serve as a \emph{shield} to
protect $\pi_{\mathbf{w}}$ at runtime under the environment context
and safety property used to synthesize $\cal{P}$.  An observation
about a current state is sent to both $\pi_{\mathbf{w}}$ and the
shield.  A high-performing neural network is allowed to take any
actions it feels are optimal as long as the next state it proposes is
still within the safety boundary formally characterized by the
inductive invariant of $\cal{C}[\cal{P}]$.  Whenever a neural policy proposes
an action that steers its controlled system out of the state space
defined by the inductive invariant we have learned as part of
deterministic program synthesis, our shield can instead take a safe
action proposed by the deterministic program $\cal{P}$.  The action
given by $\cal{P}$ is guaranteed safe because $\varphi$ defines an
inductive invariant of $\cal{C}[\cal{P}]$; taking the action allows the system to stay
within the provably safe region identified by $\varphi$.  Our
shielding mechanism is sound due to formal verification.  Because the
deterministic program was synthesized using $\pi_{\mathbf{w}}$ as an
oracle, it is expected that the shield derived from the program will
not frequently interrupt the neural network's decision, allowing the
combined system to perform (close-to) optimally.

In the inverted pendulum example, since the $90^\circ$ bound given as
a safety constraint is rather conservative, we do not expect a
well-trained neural network to violate this boundary. Indeed, 
in Fig.~\ref{fig:penduluminv}(a), even though the inductive invariant of the 
synthesized program defines a substantially smaller state space than 
what is permissible, in our simulation results, we find that the neural 
policy is never interrupted by the deterministic program when 
governing the pendulum. 
Note that the non-shaded areas in Fig.~\ref{fig:penduluminv}(a), 
while appearing safe, presumably define states from which the trajectory of 
the system can be led to an unsafe state, and would thus not be inductive.

Our synthesis approach is critical to ensuring safety when
the neural policy is used to predict actions in an environment
different from the one used during training. 
Consider a neural network suitably protected by a shield that now 
operates safely. 
The effectiveness of this shield would be greatly diminished 
if the network had to be completely retrained from scratch whenever
it was deployed in a new environment which imposes different safety
constraints.

In our running example, suppose we wish to operate the inverted
pendulum in an environment such as a Segway transporter in which the
model is prohibited from swinging significantly and whose angular
velocity must be suitably restricted.  We might specify the
following new constraints on state parameters to enforce these conditions:
\[\cal{S}_u: \{(\eta, \omega)\ \vert \neg(-30^\circ < \eta < 30^\circ \wedge
-30^\circ < \omega < 30^\circ)\}\]

\noindent Because of the dependence of a neural network to the quality
of training data used to build it, environment changes that deviate
from assumptions made at training-time could result in a costly
retraining exercise because the network must learn a new way to
penalize unsafe actions that were previously safe. However, 
training a neural network from scratch requires substantial non-trivial
effort, involving fine-grained tuning of training parameters or even
network structures.

In our framework, the existing network 
provides a reasonable approximation to the desired behavior.  To
incorporate the additional constraints defined by the new environment
$\cal{C'}$, we attempt to synthesize a new deterministic program
$\cal{P'}$ for $\cal{C'}$, a task based on our experience is substantially
easier to achieve than training a new neural network policy from
scratch.  This new program can be used to protect the original network
provided that we can use the aforementioned verification approach to
formally verify that $\cal{C'}[\cal{P'}]$ is safe by identifying a new
inductive invariant $\varphi'$.  As depicted in Fig.~\ref{fig:neuralshield}, we
simply build a new shield $\cal{S}'$ that is composed of the program
$\cal{P}'$ and the safety boundary captured by $\varphi'$.  The shield
$\cal{S}'$ can ensure the safety of the neural network in the
environment context $\cal{C}'$ with a strengthened safety condition,
despite the fact that the neural network was trained in a different
environment context $\cal{C}$.

Fig.~\ref{fig:penduluminv}(b) depicts the new unsafe states in
$\cal{C}'$ (colored in red).  It extends the unsafe range draw in
Fig.~\ref{fig:penduluminv}(a) so the inductive invariant learned there
is unsafe for the new one.  Our approach synthesizes a new
deterministic program for which we learn a more restricted inductive
invariant depicted in green in Fig.~\ref{fig:penduluminv}(b).  
{To characterize the effectiveness of our shielding approach,  we
examined 1000 simulation trajectories of this restricted version of
the inverted pendulum system, each of which is comprised of 5000 
simulation steps with the safety constraints defined by $\cal{C'}$.}
Without the shield $\cal{S}'$, the pendulum entered the unsafe region 
$\cal{S}_u$ 41 times. All of these violations were prevented by $\cal{S}'$. 
Notably, the intervention rate of $\cal{S}'$ to interrupt the neural network's
decision was extremely low.  Out of a total of 5000$\times$1000
decisions, we only observed 414 instances (0.00828\%) where the shield
interfered with (i.e., overrode) the decision made by the network.


\section{Problem Setup}
\label{sec:problem}

We model the context $\cal{C}$ of a control policy
as an environment state transition system 
$\cal{C}[\cdot] = ({X}, \cal{A}, \cal{S}, \cal{S}_0, \cal{S}_u, \cal{T}_t[\cdot], f, r)$. 
Note that $\cdot$ is intentionally left open to deploy neural control policies.
Here, ${X}$ is a finite set of variables interpreted over the reals $\mathbb{R}$
and $\cal{S} = \mathbb{R}^{X}$ is the set of all valuations of the variables ${X}$. 
We denote $s \in \cal{S}$ to be an $n$-dimensional environment state
and $a \in \cal{A}$ to be a control action where $\cal{A}$ is an infinite set of 
$m$-dimensional continuous actions that a learning-enabled controller can perform. 
We use $\cal{S}_0 \in \cal{S}$ to specify a set of initial environment states and 
$\cal{S}_u \in \cal{S}$ to specify a set of unsafe environment states that a safe
controller should avoid.
The transition relation $\cal{T}_t[\cdot]$ defines how one state transitions
to another given an action by an unknown policy.
We assume that $\cal{T}_t[\cdot]$ is governed by a 
standard {\em differential equation} $f$ defining the relationship 
between a continuously varying state $s(t)$ and action $a(t)$
and its rate of change $\dot{s}(t)$ over time $t$:
\[\dot{s}(t) = f(s(t), a(t))\]
We assume $f$ is defined by equations of the form:
$\mathbb{R}^n \times \mathbb{R}^m \rightarrow \mathbb{R}^n$
such as the example in Fig.~\ref{fig:st}. 
In the following, we often omit the time variable $t$ for simplicity.
A deterministic neural network control policy 
$\pi_{\mathbf{w}}: \mathbb{R}^n \rightarrow \mathbb{R}^m$
parameterized by a set of weight values $\mathbf{w}$ is a function 
mapping an environment state $s$ to a control action $a$ where 
\[a = \pi_{\mathbf{w}}(s)\]
By providing a feedback action, a policy can alter the 
rate of change of state variables to realize optimal system control.

The transition relation $\cal{T}_t[\cdot]$ is parameterized by 
a control policy $\pi$ that is deployed in $\cal{C}$.
We explicitly model this deployment as $\cal{T}_t[\pi]$.
Given a control policy $\pi_{\mathbf{w}}$,
we use $\cal{T}_t[\pi_{\mathbf{w}}]: \cal{S} \times \cal{S}$ to
specify all possible state transitions allowed by the policy.
We assume that a system transitions in discrete time instants $kt$
where $k = 0, 1, 2, \cdots$ and $t$ is a fixed time step ($t > 0$).  A
state $s$ transitions to a next state $s'$ after time $t$ with the
assumption that a control action $a(\tau)$ at time $\tau$ is a
constant between the time period $\tau \in [0, t)$.  Using Euler's
  method\footnote{Euler's method may sometimes poorly approximate the
    true system transition function when $f$ is highly nonlinear. More
    precise higher-order approaches such as \emph{Runge-Kutta} methods
    exist to compensate for loss of precision in this case.},
we discretize the continuous dynamics $f$ 
with finite difference approximation so it can be used in 
the discretized transition relation $\cal{T}_t[\pi_{\mathbf{w}}]$.
Then $\cal{T}_t[\pi_{\mathbf{w}}]$ can compute these estimates by 
the following \emph{difference equation}:
\[\cal{T}_t[\pi_{\mathbf{w}}] := \{(s, s')\ \vert\ s' = s + f(s, \pi_{\mathbf{w}}(s)) \times t\}\]

\noindent{\bf Environment Disturbance}. Our model allows bounded
external properties (\emph{e.g.}, additional environment-imposed constraints)
by extending the definition of change of rate: $\dot{s}= f(s, a) + d$
where $d$ is a vector of random disturbances.  We use $d$ to encode
environment disturbances in terms of bounded nondeterministic values.  
We assume that tight upper and lower bounds of $d$ can be
accurately estimated at runtime using multivariate normal distribution
fitting methods.

\noindent{\bf Trajectory.} A trajectory $h$ of a state transition system 
$\cal{C}[\pi_{\mathbf{w}}]$ which we denote as $h \in \cal{C}[\pi_{\mathbf{w}}]$
is a sequence of states $s_0, \cdots, s_{i}, s_{i+1}, \cdots$
where $s_0 \in \cal{S}_0$ and $(s_{i}, s_{i+1}) \in \cal{T}_t[\pi_{\mathbf{w}}]$
for all $i \ge 0$. We use $C \subseteq \cal{C}[\pi_{\mathbf{w}}]$ to denote a set of trajectories.
The reward that a control policy receives on performing an action $a$ in
a state $s$ is given by the reward function $r(s, a)$.

\noindent{\bf Reinforcement Learning.}
The goal of reinforcement learning is to maximize the reward that can
be collected by a neural control policy in a given environment context
$\cal{C}$.  Such problems can be abstractly formulated as
\begin{equation}\label{eq:3}
\begin{split}
\max_{\mathbf{w} \in \mathbb{R}^n} &J(\mathbf{w}) \\
J(\mathbf{w}) = \mathbb{E} &[r(\pi_{\mathbf{w}})]
\end{split}
\end{equation}
Assume that $s_0, s_1, \ldots, s_T$ is a trajectory of length $T$ of the
state transition system and $r(\pi_{\mathbf{w}}) = \lim\limits_{T \to
  \infty}\Sigma^T_{k=0}r(s_k, \pi_{\mathbf{w}}(s_k))$ is the
cumulative reward achieved by the policy $\pi_{\mathbf{w}}$ from this
trajectory.  Thus this formulation uses simulations of the transition
system with finite length rollouts to estimate the expected cumulative
reward collected over $T$ time steps and aim to maximize this reward.
{Reinforcement learning assumes polices are expressible in some {executable} structure 
(\emph{e.g.} a neural network) and allows samples generated from one policy to influence 
the estimates made for others. The two main approaches for reinforcement learning are 
value function estimation and direct policy search. 
We refer readers to~\cite{DRLOverview} for an overview.
}

\noindent{\bf Safety Verification.}
Since reinforcement learning only considers finite length rollouts, we
wish to determine if a control policy is safe to use under an
infinite time horizon.  Given a state transition system, the safety
verification problem is concerned with verifying that no trajectories
contained in $\cal{S}$ starting from an initial state in $\cal{S}_0$
reach an unsafe state in $\cal{S}_u$.  However, a neural network is a
representative of a class of deep and sophisticated models that
challenges the capability of the state-of-the-art verification
techniques.  This level of complexity is exacerbated in our work
because we consider the long term safety of a neural policy deployed
within a nontrivial environment context $\cal{C}$ that in turn is
described by a complex infinite-state transition system.

\section{Verification Procedure}
\label{sec:verificationprocedure}



To verify a neural network control policy $\pi_{\mathbf{w}}$ with
respect to an environment context $\cal{C}$, we first synthesize a
\emph{deterministic policy program} $\cal{P}$ from the neural policy.
We require that $\cal{P}$ both (\emph{a}) broadly resembles its neural oracle
and (\emph{b}) additionally satisfies a desired safety property when it is
deployed in $\cal{C}$. We conjecture that a safety proof of
$\cal{C}[\cal{P}]$ is easier to construct than that of
$\cal{C}[\pi_{\mathbf{w}}]$.  More importantly, we leverage the 
safety proof of $\cal{C}[\cal{P}]$ to ensure the safety of
$\cal{C}[\pi_{\mathbf{w}}]$.

\subsection{Synthesis}
\label{sec:synthesize}

Fig.~\ref{fig:syntax} defines a search space for a deterministic
policy program $\cal{P}$ where $E$ and $\varphi$ represent the basic
syntax of (polynomial) program expressions and inductive invariants,
respectively.  Here $v$ ranges over a universe of numerical constants,
$x$ represents variables, and $\oplus$ is a basic operator including $+$
and $\times$. A deterministic program $\cal{P}$ also features conditional
statements using $\varphi$ as branching predicates.

We allow the user to define a sketch~\cite{sketchthesis,sketch} to describe the
shape of target policy programs using the grammar in
Fig.~\ref{fig:syntax}. We use $\cal{P}[\theta]$ to represent a sketch
where $\theta$ represents unknowns that need to be filled-in by the
synthesis procedure. We use $\cal{P}_\theta$ to represent a
synthesized program with known values of $\theta$. Similarly, the user
can define a sketch of an inductive invariant $\varphi[\cdot]$ that is
used (in Sec.~\ref{sec:verify}) to learn a safety proof to verify a
synthesized program $\cal{P}_\theta$.

\begin{figure}[t]
\begin{center}
\begin{align}
E &::= v\ \vert\ x\ \vert\ \oplus(E_1, \ldots, E_k) \nonumber\\
\varphi &::= E \le 0 \nonumber \nonumber\\
\cal{P} &::= \text{\bf return }\ E\ \vert\ \text{\bf if}\ \varphi\ \text{\bf then}\ \text{\bf return}\ E\ \text{\bf else}\ \cal{P} \nonumber
\end{align}
\caption{Syntax of the Policy Programming Language.}
\label{fig:syntax}
\end{center}
\end{figure}

We do not require the user to explicitly define conditional statements
in a program sketch. Our synthesis algorithm uses verification
counterexamples to lazily add branch predicates $\varphi$ under which
a program performs different computations depending on 
whether $\varphi$ evaluates to true or false.
The end user simply writes a sketch over basic expressions. For example,
a sketch that defines a family of linear function over a collection of variables can
be expressed as:
\begin{equation}\label{eq:linearsketch}
{\cal P}[\theta](X) ::= \text{\bf return}\ \theta_1 x_1 + \theta_2 x_2 + \cdots + \theta_n x_n + \theta_{n+1}
\end{equation}
Here $X = (x_1, x_2, \cdots)$ are system variables in which
the coefficient constants $\theta = (\theta_1, \theta_2, \cdots)$ are unknown.

The goal of our synthesis algorithm is to find unknown values of 
$\theta$ that maximize the likelihood that $\cal{P}_\theta$ 
resembles the neural oracle $\pi_\mathbf{w}$ while still being a 
safe program with respect to the environment context $\cal{C}$:
\begin{equation}\label{eq:randomsearch}
\theta^* = \argmaxA_{\theta \in \mathbb{R}^{n+1}} d(\pi_{\mathbf{w}}, \cal{P}_\theta, \cal{C})
\end{equation}
where $d$ measures the \emph{distance} between the outputs of the
estimate program $\cal{P}_\theta$ and the neural policy
$\pi_{\mathbf{w}}$ subject to safety constraints.  
To avoid the computational complexity that arises if we consider a
solution to this equation analytically as an optimization problem, we
instead approximate $d(\pi_{\mathbf{w}}, \cal{P}_\theta, \cal{C})$ by
randomly sampling a set of trajectories $C$ that are encountered by
$\cal{P}_\theta$ in the environment state transition system
$\cal{C}[\cal{P}_\theta]$:
\[d(\pi_{\mathbf{w}}, \cal{P}_\theta, \cal{C}) \approx d(\pi_{\mathbf{w}}, \cal{P}_\theta, C)\
s.t.\ C \subseteq \cal{C}[\cal{P}_\theta]\]
We estimate $\theta^*$ using these sampled trajectories $C$ and define:
\[
d(\pi_{\mathbf{w}}, \cal{P}_\theta, C) = \sum_{h \in C} d(\pi_{\mathbf{w}}, \cal{P}_\theta, h)
\]
Since each trajectory $h \in C$ is a finite rollout $s_0, \ldots, s_T$ of length $T$, we have:
\[d(\pi_{\mathbf{w}}, \cal{P}_\theta, h) = \sum_{t=0}^T
\begin{cases}-\norm{(\cal{P}_\theta(s_t) - \pi_{\mathbf{w}}(s_t))} & s_t \notin \cal{S}_u
\\ -MAX & s_t \in \cal{S}_u \end{cases}\]

\noindent where $\norm{\cdot}$ is a suitable norm. As illustrated in Sec.~\ref{sec:safetyverification},
we aim to minimize the distance between a synthesized program $\cal{P}_\theta$ 
from a sketch space and its neural oracle along sampled trajectories encountered 
by $\cal{P}_\theta$ in the environment context but put a large penalty on states that are unsafe.

\SetKwRepeat{Do}{do}{while}
\RestyleAlgo{boxruled}
\IncMargin{.7em}
\begin{algorithm}[!t]
  \small
\caption{Synthesize ($\pi_\mathbf{w}$, $\cal{P}[\theta]$, $\cal{C}[\cdot]$)}
\label{policyextraction}
$\theta$ $\leftarrow$ $\mathbf{0}$\;
\Do{$\theta$ is not converged}{
Sample $\delta$ from a zero mean Gaussian vector\;
Sample a set of trajectories $C_1$ using $\cal{C}[\cal{P}_{\theta+\nu\delta}]$\;
Sample a set of trajectories $C_2$ using $\cal{C}[\cal{P}_{\theta-\nu\delta}]$\;
$\theta \leftarrow \theta + {\alpha}[\frac{d(\pi_{\mathbf{w}}, \cal{P}_{\theta+\nu\delta}, C_1) - d(\pi_{\mathbf{w}}, \cal{P}_{\theta-\nu\delta}, C_2)}{\nu}]\delta$\;
}
\Return{$\cal{P}_\theta$} 
\end{algorithm}
\DecMargin{.7em}

\noindent{\bf Random Search.}
We implement the idea encoded in equation~\eqref{eq:randomsearch} in
Algorithm~\ref{policyextraction} that depicts the pseudocode of 
our policy interpretation approach. It take as inputs a neural policy, 
a policy program sketch parameterized by $\theta$, and an environment
state transition system and outputs a synthesized policy 
program $\cal{P}_\theta$.

An efficient way to solve equation~\eqref{eq:randomsearch} is to directly perturb 
$\theta$ in the search space by adding random noise and then update $\theta$ 
based on the effect on this perturbation~\cite{randomsearchbasic}. We choose 
a direction uniformly at random 
on the sphere in parameter space, and then optimize the goal function along this direction.
To this end, in line 3 of Algorithm~\ref{policyextraction}, we sample
Gaussian noise $\delta$ to be added to policy parameters $\theta$ in both directions
where $\nu$ is a small positive real number. 
In line 4 and line 5, we sample trajectories $C_1$ and $C_2$ from the 
state transition systems obtained by running the perturbed policy program 
$\cal{P}_{\theta + \nu\delta}$ and $\cal{P}_{\theta - \nu\delta}$ 
in the environment context $\cal{C}$.

We evaluate the proximity of these two policy programs to the neural oracle
and, in line 6 of Algorithm~\ref{policyextraction}, to improve the policy program, 
we also optimize equation~\eqref{eq:randomsearch} by updating $\theta$ with a finite 
difference approximation along the direction:
\begin{equation}\label{randomsearch}
\theta \leftarrow \theta + {\alpha}[\frac{d(\cal{P}_{\theta+\nu\delta}, \pi_{\mathbf{w}}, C_1) - d(\cal{P}_{\theta-\nu\delta}, \pi_{\mathbf{w}}, C_2)}{\nu}]\delta
\end{equation}
where $\alpha$ is a predefined learning rate.
Such an update increment corresponds
to an unbiased estimator of the gradient of $\theta$~\cite{randomsearchbasic2,NIPS2018} 
for equation~\eqref{eq:randomsearch}. The algorithm iteratively updates 
$\theta$ until convergence.

\subsection{Verification}
\label{sec:verify}

A synthesized policy program $\cal{P}$ is verified with respect to
an environment context given as an infinite state transition system defined 
in Sec.~\ref{sec:problem}:
$\cal{C}[\cal{P}] = ({X}, \cal{A}, \cal{S}, \cal{S}_0, \cal{S}_u, \cal{T}_t[\cal{P}], f, r)$.
Our verification algorithm learns an inductive invariant
$\varphi$ over the transition relation $\cal{T}_t[\cal{P}]$ formally proving
that all possible system behaviors are encapsulated in $\varphi$
and $\varphi$ is required to be disjoint with all unsafe
states $\cal{S}_u$.

We follow template-based constraint solving approaches for inductive
invariant
inference~\cite{template1,template2,barriercertificate,huikong} to
discover this invariant.  The basic idea is to identify a function $E:
\mathbb{R}^n \rightarrow \mathbb{R}$ that serves as a "barrier"~\cite{template1,barriercertificate,huikong}
between reachable system states (evaluated to be nonpositive by $E$), and
unsafe states (evaluated positive by $E$).  
Using the invariant syntax in Fig.~\ref{fig:syntax}, the user can define an invariant sketch 
\begin{equation}\label{eq:invariantsketch}
\varphi[c](X) ::= E[c](X) \le 0
\end{equation} 
over variables $X$ and $c$ unknown coefficients intended to be synthesized. 
Fig.~\ref{fig:syntax} carefully restricts that an invariant sketch $E[c]$ 
can only be postulated as a polynomial function as there exist
efficient SMT solvers~\cite{z3} and constraint solvers~\cite{sos} 
for nonlinear polynomial reals. 
Formally, assume real coefficients $c = (c_0, \cdots, c_p)$ 
are used to parameterize $E[c]$ in an affine manner:
\[E[c](X) = \Sigma^p_{i=0}c_i b_i(X)\] 
where the $b_i(X)$'s are some monomials in variables $X$.  As a
heuristic, the user can simply determine an upper bound on the degree
of $E[c]$, and then include all monomials whose degrees are no greater
than the bound in the sketch.  Large values of the bound enable
verification of more complex safety conditions, but impose greater
demands on the constraint solver; small values capture coarser safety
properties, but are easier to solve.  

\begin{example}[label=ex1]
Consider the inverted pendulum system in Sec.~\ref{sec:safetyverification}.
To discover an inductive invariant for the transition system, the user might
choose to define an upper bound of 4, which results in the following
polynomial invariant sketch:
$\varphi[c](\eta, \omega) ::= E[c](\eta, \omega) \le 0$ where
\[E[c](\eta, \omega) = c_0\eta^4 + c_1\eta^3\omega + c_2\eta^2\omega^2 + c_3\eta\omega^3 + c_4\omega^4 + c_5\eta^3 + \cdots + c_p\]
The sketch includes all monomials over $\eta$ and $\omega$,
whose degrees are no greater than 4. The coefficients $c = [c_0, \cdots, c_p]$ 
are unknown and need to be synthesized.
\end{example}

To synthesize these unknowns, we require that $E[c]$ must satisfy the 
following verification conditions:
\begin{align}
& \forall(s) \in \cal{S}_u & & E[c](s) > 0  \label{unsafe} \\
& \forall(s) \in \cal{S}_0 & & E[c](s) \le 0 \label{init} \\
& \forall(s, s') \in \cal{T}_t[\cal{P}] & & E[c](s') - E[c](s) \le 0. \label{induction}
\end{align} 
We claim that $\varphi::= E[c](x) \le 0$ defines an inductive invariant 
because verification condition~\eqref{init} ensures that any initial state
$s_0 \in S_0$ satisfies $\varphi$ since $E[c](s_0) \le 0$;
verification condition~\eqref{induction} asserts that along the transition 
from a state $s \in \varphi$ (so $E[c](s) \le 0$) to a resulting state $s'$, 
$E[c]$ cannot become positive so $s'$ satisfies $\varphi$ as well.  
Finally, according to verification condition~\eqref{unsafe}, 
$\varphi$ does not include any unsafe state 
$s_u \in \cal{S}_u$ as $E[c](s_u)$ is positive.

Verification conditions~\eqref{unsafe}~\eqref{init}~\eqref{induction}
are polynomials over reals. Synthesizing unknown coefficients can be left to 
an SMT solver~\cite{z3} after universal quantifiers are eliminated using a 
variant of Farkas Lemma as in~\cite{template1}. 
However, observe that $E[c]$ is convex.\footnote{For arbitrary 
$E_1(x)$ and $E_2(x)$ satisfying the verification conditions and $\alpha \in [0,1]$,
$E(x) = \alpha E_1(x) + (1-\alpha)E_2(x)$ satisfies the conditions as well.}
We can gain efficiency by finding unknown coefficients $c$ using off-the-shelf 
convex constraint solvers following~\cite{barriercertificate}.
Encoding verification conditions~\eqref{unsafe}~\eqref{init}~\eqref{induction}
as polynomial inequalities, we search $c$ that can prove non-negativity of 
these constraints via an efficient and convex sum of squares programming solver~\cite{sos}.
Additional technical details are provided in the supplementary material~\cite{techreport}.

\noindent{\bf Counterexample-guided Inductive Synthesis (CEGIS)}. 
Given the environment state transition system $\cal{C}[\cal{P}]$ deployed with
a synthesized program $\cal{P}$, the verification approach
above can compute an inductive invariant over the state transition system
or tell if there is no feasible solution in the given set of candidates.
Note however that the latter does not necessarily imply that the system is unsafe.

Since our goal is to learn a safe deterministic program from a neural
network, we develop a counterexample guided inductive program
synthesis approach.  A CEGIS algorithm in our context is challenging
because safety verification is necessarily incomplete, and may not be
able to produce a counterexample that serves as an explanation for why
a verification attempt is unsuccessful.

We solve the incompleteness challenge by leveraging the fact that we
can simultaneously synthesize and verify a program. Our CEGIS approach 
is given in Algorithm~\ref{learnverifiedcontroller}. A counterexample is an 
initial state on which our synthesized program is not yet proved safe.
Driven by these counterexamples, our algorithm synthesizes a set of programs 
from a sketch along with the conditions under which we can switch from
from one synthesized program to another.

 \SetKwRepeat{Do}{do}{while}
 \RestyleAlgo{boxruled}
 \IncMargin{.7em}
 \begin{algorithm}[!t]
   \small
 \caption{CEGIS ($\pi_{\mathbf{w}}$, $\cal{P}[\theta]$, $\cal{C}[\cdot]$)}
 \label{learnverifiedcontroller}
 {\sf policies} $\leftarrow$ $\emptyset$\;
 {\sf covers} $\leftarrow$ $\emptyset$\;
 \While{$\cal{C}.\cal{S}_0$ $\not\subseteq$ {\sf covers}}{
  search $s_0$ such that $s_0 \in \cal{C}.\cal{S}_0 \wedge s_0 \notin {\sf covers}$\; 
  $r^*$ $\leftarrow$ Diameter($\cal{C}.\cal{S}_0$)\;
  \Do{True}{
    $\phi_{bound}$ $\leftarrow$ $\{s\ \vert\ s \in \{s_0 - r^*, s_0 + r^*\}\}$\;
    $\cal{\tilde C}$ $\leftarrow$ $\cal{C}$ {\bf where} $\cal{\tilde C}.\cal{S}_0 = (\cal{C}.\cal{S}_0 \cap \phi_{bound})$\;
    $\theta$ $\leftarrow$ Synthesize($\pi_{\mathbf{w}}$,  $\cal{P}[\theta]$, $\cal{\tilde C}[\cdot]$)\; 
    $\varphi$ $\leftarrow$ Verify($\cal{\tilde C}[\cal{P}_\theta]$)\; 
    \uIf{$\varphi$ is False}{
      $r^*$ $\leftarrow$ $r^*/2$\;
    } \uElse{
      {\sf covers} $\leftarrow$ {\sf covers} $\cup$ $\{s\ \vert\ \varphi(s)\}$\;
      {\sf policies} $\leftarrow$ {\sf policies} $\cup$ ($\cal{P}_\theta$, $\varphi$)\;
      \bf{break}\;
    }
  }
 }
 \Return{{\sf policies}}
 \end{algorithm}
 \DecMargin{.7em}

Algorithm~\ref{learnverifiedcontroller} takes as input a neural policy 
$\pi_{\mathbf{w}}$, a program sketch $\cal{P}[\theta]$ and 
an environment context $\cal{C}$. It maintains synthesized policy
programs in {\sf policies} in line 1, each of which is inductively verified
safe in a partition of the universe state space that is maintained in 
{\sf covers} in line 2. 
For soundness, the state space covered by such partitions must be 
able to include all initial states, checked in line 3 of 
Algorithm~\ref{learnverifiedcontroller} by an SMT solver.
In the algorithm, we use $\cal{C}.\cal{S}_0$ to access a field of $\cal{C}$ 
such as its initial state space.

The algorithm iteratively samples a counterexample initial state $s_0$ 
that is currently not covered by {\sf covers} in line 4. 
Since {\sf covers} is empty at the beginning, this choice is 
uniformly random initially;
we synthesize a presumably safe policy program in line 9 of
Algorithm~\ref{learnverifiedcontroller} that resembles the neural
policy $\pi_{\mathbf{w}}$ considering all possible initial states $\cal{S}_0$
of the given environment model $\cal{C}$, using Algorithm~\ref{policyextraction}.

If verification fails in line 11, our approach simply reduces the
initial state space, hoping that a safe policy program is easier to
synthesize if only a subset of initial states are considered.  The
algorithm in line 12 gradually shrinks the radius $r^*$ of the initial
state space around the sampled initial state $s_0$.  The synthesis
algorithm in the next iteration synthesizes and verifies a candidate
using the reduced initial state space.  The main idea is that if the
initial state space is shrunk to a restricted area around $s_0$ but a
safety policy program still cannot be found, it is quite possible that
either $s_0$ points to an unsafe initial state of the neural oracle or
the sketch is not sufficiently expressive.

If verification at this stage succeeds with an inductive invariant
$\varphi$,  a new policy program $\cal{P}_\theta$ is synthesized
that can be verified safe in the state space covered by $\varphi$. 
We add the inductive invariant and the policy program
into {\sf covers} and {\sf policies} in line 14 and 15 respectively 
and then move to another iteration of counterexample-guided
synthesis. This iterative \emph{synthesize-and-verify} process 
continues until the entire initial state space is covered (line 3 to 18).
The output of Algorithm~\ref{learnverifiedcontroller} is 
$[({\cal{P}_\theta}_1, \varphi_1), ({\cal{P}_\theta}_2, \varphi_2), \cdots]$
that essentially defines conditional statements in a synthesized
program performing different actions depending on whether a
specified invariant condition evaluates to true or false.

\begin{theorem}
  If $CEGIS(\pi_w, \cal{P}[\theta], \cal{C}[\cdot]) = [({\cal{P}_\theta}_1, \varphi_1), ({\cal{P}_\theta}_2,\\ \varphi_2), \cdots]$
  (as defined in Algorithm~\ref{learnverifiedcontroller}), then the deterministic program $\cal{P}$:
\begin{center}
\begin{python}[mathescape=true]
$\lambda X$.if $\varphi_1(X)$: return ${\cal{P}_\theta}_1(X)$ else if $\varphi_2(X)$: return ${\cal{P}_\theta}_2(X)$ $\cdots$
\end{python}
\end{center}
is safe in the environment $\cal{C}$ meaning that $\varphi_1 \vee \varphi_2 \vee \cdots$ is an inductive invariant
of $\cal{C}[\cal{P}]$ proving that $\cal{C}.\cal{S}_u$ is unreachable.
\end{theorem}

\begin{figure}[t]
\centering
\includegraphics[width=0.47\textwidth]{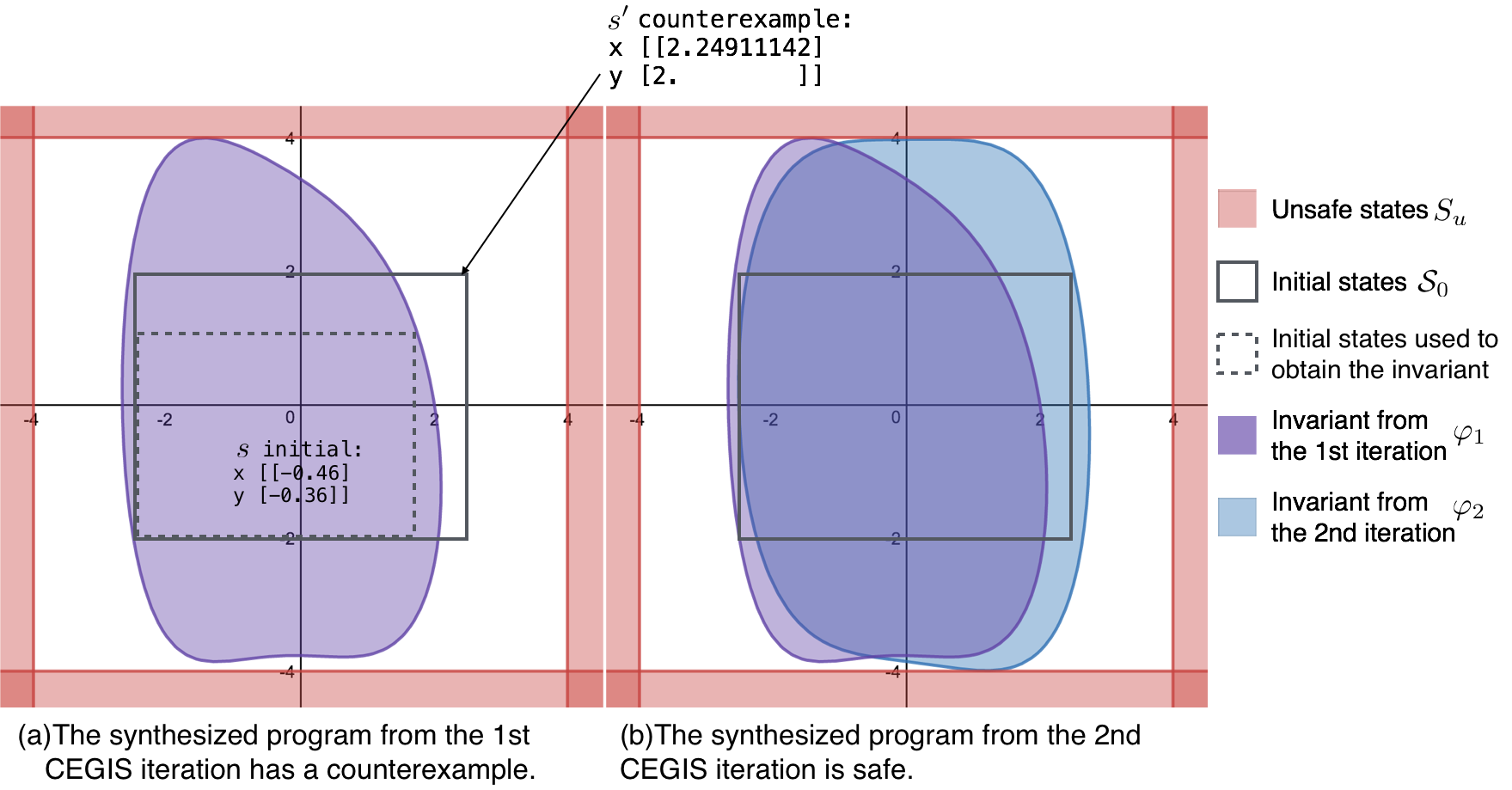}
\caption{CEGIS for Verifiable Reinforcement Learning.} 
\label{fig:cegisexample}
\end{figure}

\begin{example}
We illustrate the proposed counterexample guided inductive synthesis method 
by means of a simple example,
the Duffing oscillator~\cite{oscillator}, a nonlinear second-order environment.
The transition relation of the environment system $\cal{C}$ is described 
with the differential equation:
 \begin{align}
 &\dot{x} = y \nonumber \\
 &\dot{y} = -0.6y - x - x^3 + a \nonumber
 \end{align}
where $x, y$ are the state variables and $a$ the continuous control action
given by a well-trained neural feedback control policy $\pi$ such that $a = \pi(x, y)$.
The control objective is to regulate the state to the origin from a set of initial states
$\cal{C}.\cal{S}_0: \{x, y \vert -2.5 \le x \le 2.5 \wedge -2 \le y \le 2\}$.
To be safe, the controller must be able to avoid a set of 
unsafe states $\cal{C}.\cal{S}_u: \{x, y \vert\ \neg (-5 \le x \le 5 \wedge -5 \le y \le 5)\}$.
Given a  program sketch as in the pendulum example, that is
$\cal{P}[\theta](x, y) ::= \theta_1 x + \theta_2 y$, the user can ask
the constraint solver to reason over a small (say 4) order polynomial invariant sketch 
for a synthesized program as in Example~\ref{ex1}.

Algorithm~\ref{learnverifiedcontroller} initially samples an
initial state $s$ as $\{x = -0.46, y = -0.36\}$ from $\cal{S}_0$. 
The inner do-while loop of the algorithm can discover a sub-region
of initial states in the dotted box of Fig.~\ref{fig:cegisexample}(a)
that can be leveraged to synthesize a verified deterministic policy
$\cal{P}_1(x, y) ::= 0.39x -1.41y$ from the sketch. We also obtain 
an inductive invariant showing that the synthesized policy can 
always maintain the controller within the invariant set drawn
in purple in Fig.~\ref{fig:cegisexample}(a):
$
\varphi_1 \equiv 20.9x^4 + 2.9x^3y + 1.4x^2y^2 + 0.4xy^3 + 29.6x^3 + 20.1x^2y + 11.3xy^2 + 1.6y^3 + 25.2x^2 + 39.2xy + 53.7y^2 - 680 \le 0
$.

This invariant explains why the initial state space used for verification
does not include the entire $\cal{C}.\cal{S}_0$: a counterexample initial
state $s' = \{x = 2.249, y = 2\}$ is not covered by the invariant
for which the synthesized policy program above is not verified safe.
The CEGIS loop in line 3 of Algorithm~\ref{learnverifiedcontroller} 
uses $s'$ to synthesize another
deterministic policy $\cal{P}_2(x,y) ::= 0.88x -2.34y$ from the sketch
whose learned inductive invariant is depicted in blue in Fig.~\ref{learnverifiedcontroller}(b):
$
\varphi_2 \equiv 12.8x^4 + 0.9x^3y - 0.2x^2y^2 - 5.9x^3 - 1.5xy^2 - 0.3y^3 + 2.2x^2 + 4.7xy + 40.4y^2 - 619 \le 0
$.
Algorithm~\ref{learnverifiedcontroller} then terminates because 
$\varphi_1 \vee \varphi_2$ covers $\cal{C}.\cal{S}_0$.
Our system interprets the two synthesized deterministic policies 
as the following deterministic program $\cal{P}_{\mathit{oscillator}}$ using
the syntax in Fig.~\ref{fig:syntax}:
\begin{center}
\begin{python}[mathescape=true]
def $\cal{P}_{\mathit{oscillator}}\ (x, y)$:
  if $20.9x^4 + 2.9x^3y + 1.4x^2y^2 + \cdots + 53.7y^2 - 680 \le 0$: # $\color{red} \varphi_1$
    return $0.39x -1.41y$
  else if $12.8x^4 + 0.9x^3y - 0.2x^2y^2 + \cdots + 40.4y^2 - 619 \le 0$: # $\color{red} \varphi_2$
    return $0.88x -2.34y$
  else abort # unsafe
\end{python}
\end{center}
Neither of the two deterministic policies enforce safety by themselves on all initial
states but do guarantee safety when combined together because
by construction, $\varphi = \varphi_1 \vee \varphi_2$
is an inductive invariant of $\cal{P}_{\mathit{oscillator}}$ in the 
environment $\cal{C}$.
\end{example}

{Although Algorithm~\ref{learnverifiedcontroller} is sound, it may not 
terminate as $r^*$ in the algorithm can become arbitrarily small and 
there is also no restriction on the size of potential counterexamples.
Nonetheless, our experimental results indicate that the algorithm 
performs well in practice.}

\subsection{Shielding}
\label{sec:shield}

A safety proof of a synthesized deterministic program of 
a neural network does not automatically lift to a safety argument 
of the neural network from which it was derived since the network 
may exhibit behaviors not captured by the simpler deterministic program.  
To bridge this divide, we recover soundness at runtime by
monitoring system behaviors of a neural network in its environment 
context by using the synthesized policy program and its inductive invariant
as a shield. The pseudo-code for using a shield is given in
Algorithm~\ref{shield}.  In line 1 of Algorithm~\ref{shield}, for a
current state $s$, we use the state transition system of our
environment context model to predict the next state $s'$. If $s'$ is
not within $\varphi$, we are unsure whether entering $s'$ would
inevitably make the system unsafe as we lack a proof that the neural
oracle is safe.  However, we do have a guarantee that if we follow the
synthesized program $\cal{P}$, the system would stay within the safety
boundary defined by $\varphi$ that
Algorithm~\ref{learnverifiedcontroller} has formally proved.  We do so
in line 3 of Algorithm~\ref{shield}, using $\cal{P}$ and
$\varphi$ as shields, intervening only if necessary so as to restrict
the shield from unnecessarily intervening the neural policy.
\footnote{We also extended our approach to 
synthesize deterministic programs which can guarantee
stability in the supplementary material~\cite{techreport}.}

\section{Experimental Results}
\label{sec:eval}

\begin{small}
\begin{table*}[t]
\caption{Experimental Results on Deterministic Program Synthesis, Verification, and Shielding.}
\label{tbl:result1}
\begin{tabular}{|l | c | c  c c | c c c c | c c|}
    \hline
   \rowcolor{black!20} &  &  \multicolumn{3}{c|}{Neural Network} & \multicolumn{4}{c|}{Deterministic Program as Shield} & \multicolumn{2}{c|}{Performance}  \\ 
    \rowcolor{black!20} \multirow{-2}{*}{Benchmarks} & \multirow{-2}{*}{Vars} &  Size & Training & Failures & Size & Synthesis & Overhead & Interventions & NN & Program \\ \hline
     {\sf Satellite} & 2 & $240\times200$ & 957s & 0 & 1 & 160s  & 3.37\% & 0 & 5.7 & 9.7 \\
     {\sf DCMotor} & 3 & $240\times200$ & 944s & 0 & 1 & 68s  & 2.03\% & 0 & 11.9 & 12.2 \\
     {\sf Tape} & 3 & $240\times200$ & 980s & 0 & 1  & 42s & 2.63\% & 0 & 3.0 & 3.6 \\
     {\sf Magnetic Pointer} & 3 & $240\times200$ & 992s & 0 & 1  & 85s & 2.92\% & 0 & 8.3 & 8.8  \\
     {\sf Suspension} & 4 & $240\times200$ & 960s & 0 & 1  & 41s & 8.71\% & 0 & 4.7 & 6.1 \\
     \hline
     {\sf Biology} & 3 & $240\times200$ & 978s & 0 & 1 & 168s  & 5.23\% & 0 & 2464 & 2599 \\
     {\sf DataCenter Cooling} & 3 & $240\times200$ & 968s & 0 & 1 & 168s  & 4.69\% & 0 & 14.6 & 40.1 \\
     \hline
     {\sf Quadcopter} & 2 & $300 \times 200$ & 990s & 182 & 2  & 67s & 6.41\% & 185 & 7.2 & 9.8 \\
     {\sf Pendulum} & 2 & $240\times 200$ & 962s & 60 & 3  & 1107s & 9.65\% & 65 & 44.2 & 58.6 \\
     {\sf CartPole} & 4 & $300\times200$ & 990s & 47 & 4 & 998s  & 5.62\% & 1799 & 681.3 & 1912.6 \\
     \hline
     {\sf Self-Driving} & 4 & $300\times200$ & 990s & 61 & 1  & 185s  & 4.66\% & 236 & 145.9 & 513.6 \\
     {\sf Lane Keeping} & 4 & $240\times200$ & 895s & 36 & 1  & 183s  & 8.65\% & 64 & 375.3 & 643.5 \\
     {\sf 4-Car platoon} & 8 & $500\times400\times300$ & 1160s & 8 & 4  & 609s & 3.17\% & 8 & 7.6 & 9.6 \\
     {\sf 8-Car platoon} & 16 & $500\times400\times300$ & 1165s & 40 & 1  & 1217s & 6.05\% & 1080 & 38.5 & 55.4 \\
     \hline
     {\sf Oscillator} & 18 & $240\times200$ & 1023s & 371 & 1  & 618s & 21.31\% & 93703 & 693.5 & 1135.3 \\
    \hline
  \end{tabular}
\end{table*}
\end{small}

We have applied our framework on a number of challenging control- and
cyberphysical-system benchmarks.  We consider the utility of our approach 
for verifying the safety of trained neural network controllers.  
We use the deep policy gradient algorithm~\cite{ddpg} for neural network training, 
the Z3 SMT solver~\cite{z3} to check convergence of the CEGIS loop, and the 
Mosek constraint solver~\cite{mosek} to generate inductive invariants of
synthesized programs from a sketch.  {All of our benchmarks are
verified using the program sketch defined in
equation~\eqref{eq:linearsketch} and the invariant sketch defined in
equation~\eqref{eq:invariantsketch}.}    
We report simulation results on our benchmarks 
over 1000 runs  (each run consists of 5000 simulation steps). 
Each simulation time step is fixed 0.01 second.
Our experiments were conducted on a standard desktop machine
consisting of Intel(R) Core(TM) i7-8700 CPU cores and 64GB memory.

\RestyleAlgo{boxruled}
\IncMargin{.7em}
\begin{algorithm}[!t]
  \small
\caption{Shield ($s$, $\cal{C}[\pi_{\mathbf{w}}]$, $\cal{P}$, $\varphi$)}
\label{shield}
Predict $s'$ such that $(s, s') \in \cal{C}[\pi_{\mathbf{w}}].\cal{T}_t(s)$\;
\lIf{$\varphi(s')$}{
  \Return{$\pi_{\mathbf{w}}(s)$}
}
\lElse{
  \Return{$\cal{P}(s)$}
}
\end{algorithm}
\DecMargin{.7em}

{\paragraph{Case Study on Inverted Pendulum.} 
We first give more details on the evaluation result of our running example,
the inverted pendulum. Here we consider a more restricted safety condition that
deems the system to be unsafe when the pendulum's angle is more than 23$^\circ$ from
the origin (\emph{i.e.,} significant swings are further prohibited).
The controller is a 2 hidden-layer neural model ($240 \times 200$).
Our tool interprets the neural network as a program
containing three conditional branches:
\begin{center}
\begin{python}[mathescape=true]
def $\cal{P}\ (\eta, \omega)$:
  if $17533\eta^4 + 13732\eta^3\omega + 3831\eta^2\omega^2 - 5472\eta\omega^3 + 8579\omega^4 + 6813\eta^3 +$ 
     $9634\eta^2\omega + 3947\eta\omega^2 - 120\omega^3 + 1928\eta^2 + 1915\eta\omega + 1104\omega^2 - 313 \le 0$:
    return $-17.28176866\eta -10.09441768\omega$
  else if $2485\eta^4 + 826\eta^3\omega - 351\eta^2\omega^2 + 581\eta\omega^3 + 2579\omega^4 + 591\eta^3 +$ 
            $9\eta^2\omega + 243\eta\omega^2 - 189\omega^3 + 484\eta^2 + 170\eta\omega + 287\omega^2 - 82 \le 0$:
    return $-17.34281984x -10.73944835y$
  else if $115496\eta^4 + 64763\eta^3\omega + 85376\eta^2\omega^2 + 21365\eta\omega^3 + 7661\omega^4 -$
          $111271\eta^3 - 54416\eta^2\omega - 66684\eta\omega^2 - 8742\omega^3 + 33701\eta^2 +$ 
          $11736\eta\omega + 12503\omega^2 - 1185 \le 0$:  
    return $-25.78835525\eta -16.25056971\omega$
  else abort
\end{python}
\end{center}
Over 1000 simulation runs, we found 60 unsafe cases when running the 
neural controller alone. Importantly, running the neural controller in tandem 
with the above verified and synthesized program can prevent all  unsafe 
neural decisions. There were only with 65 interventions from the 
program on the neural network.  Our results demonstrate that CEGIS is 
important to ensure a synthesized program is safe to use. We report more 
details including synthesis times below. 
}

{One may ask why we do not directly learn a deterministic program
to control the device (without appealing to the neural policy at all), using reinforcement 
learning to synthesize its unknown parameters. Even for an example as simple as 
the inverted pendulum, our discussion above shows that it is difficult (if not impossible) to 
derive a single straight-line (linear) program that is safe to control 
the system. Even for scenarios in which a straight-line program suffices,
using existing RL methods to directly learn unknown parameters in our sketches may
still fail to obtain a safe program.
For example, we considered if a linear control policy can be learned to (1) prevent
the pendulum from falling down (2) and require a pendulum control action
to be strictly within the range $[-1, 1]$ (\emph{e.g.}, operating the controller in an 
environment with low power constraints).  We found that despite many experiments 
on tuning learning rates and rewards, directly training a linear control program
to conform to this restriction with either reinforcement learning (e.g. policy gradient) 
or random search~\cite{NIPS2018} was unsuccessful because of undesirable
overfitting.  In contrast, neural networks work much better for these 
RL algorithms and can be used to guide the synthesis of a deterministic 
program policy. 
Indeed, by treating the neural policy as an oracle, we were able to quickly 
discover a straight-line linear deterministic program that in fact satisfies this additional 
motion constraint.}

\paragraph{Safety Verification.} Our verification results are given in
Table~\ref{tbl:result1}. In the table, {\bf Vars} represents the
number of variables in a control system - this number serves as a
proxy for application complexity; {\bf Size} the number of neurons in
hidden layers of the network; the {\bf Training} time for the network;
and, its {\bf Failures}, the number of times the network failed to
satisfy the safety property in simulation. The table also gives the {\bf
  Size} of a synthesized program in term of the number of polices
found by Algorithm~\ref{learnverifiedcontroller} (used to generate
conditional statements in the program); its {\bf Synthesis} time; the
{\bf Overhead} of our approach in terms of the additional cost
(compared to the non-shielded variant) in running time to use a
shield; and, the number of {\bf Interventions}, the number of
times the shield was invoked across all simulation runs.
{We also report performance gap between of a shield neural 
policy ({\bf NN}) and a purely programmatic policy ({\bf Program}), 
in terms of the number of steps on average that a controlled system 
spends in reaching a steady state of the system (\emph{i.e.,} 
a convergence state).
}

The first five benchmarks are linear time-invariant control systems
adapted from~\cite{realsyn}.  The safety property is that the reach
set has to be within a safe rectangle.  Benchmark {\sf Biology}
defines a minimal model of glycemic control in diabetic patients such
that the dynamics of glucose and insulin interaction in the blood
system are defined by polynomials~\cite{biology}.  For safety, we
verify that the neural controller ensures that the level of plasma
glucose concentration is above a certain threshold. Benchmark {\sf
  DataCenter Cooling} is a model of a collection of three server
racks each with their own cooling devices and they also shed heat to
their neighbors. The safety property is that a learned controller must
keep the data center below a certain temperature.  In these benchmarks, the cost
to query the network oracle constitutes the dominant time for
generating the safety shield in these benchmarks.  Given the
simplicity of these benchmarks, the neural network controllers did not
violate the safety condition in our trials, and moreover there were no
interventions from the safety shields that affected performance.


The next three benchmarks {\sf Quadcopter}, {\sf (Inverted) Pendulum}
and {\sf Cartpole} are selected from classic control applications and
have more sophisticated safety conditions.  We have discussed the
inverted pendulum example at length earlier. The {\sf Quadcopter}
environment tests whether a controlled quadcopter can realize stable
flight. The environment of {\sf Cartpole} consists of a pole attached
to an unactuated joint connected to a cart that moves along a
frictionless track.  The system is unsafe when the pole's angle is
more than 30$^\circ$ from being upright or the cart moves by more than
0.3 meters from the origin.  We observed safety violations in each of
these benchmarks that were eliminated using our verification
methodology.  Notably, the number of interventions is remarkably low,
as a percentage of the overall number of simulation steps.

Benchmark {\sf Self-driving} defines a single car navigation problem. The 
neural controller is responsible for preventing the car from veering into 
canals found on either side of the road.
Benchmark {\sf Lane Keeping} models another safety-related car-driving
problem. The neural controller aims to maintain a vehicle between 
lane markers and keep it centered in a possibly curved lane.
The curvature of the road is considered as a disturbance input.
{Environment disturbances of such kind can be conservatively specified 
in our model, accounting for noise and unmodeled dynamics.
Our verification approach supports these disturbances 
(verification condition~\eqref{induction}).}
Benchmarks $n$-{\sf Car platoon} model multiple ($n$) vehicles 
forming a platoon, maintaining a safe relative distance among one another~\cite{platoon}.
Each of these benchmarks exhibited some number of violations that were remediated
by our verification methodology.
Benchmark {\sf Oscillator} consists of a two-dimensional switched
oscillator plus a 16-order filter. The filter smoothens the
input signals and has a single output signal. We verify that
the output signal is below a safe threshold.
Because of the model complexity of this benchmark, it exhibited
significantly more violations than the others.  Indeed, the
neural-network controlled system often oscillated between the safe and
unsafe boundary in many runs.  Consequently, the overhead in this
benchmark is high because a large number of shield interventions was
required to ensure safety.  In other words, the synthesized shield
trades performance for safety to guarantee that the threshold boundary
is never violated.

For all benchmarks, our tool successfully generated safe interpretable
deterministic programs and inductive invariants as shields.  When a neural
controller takes an unsafe action, the synthesized shield correctly
prevents this action from executing by providing an alternative
provable safe action proposed by the verified deterministic program.
{In term of performance, Table~\ref{tbl:result1} shows that a shielded 
neural policy is a feasible approach to drive a controlled system into 
a steady state. For each of the benchmarks studied, the
programmatic policy is less performant than the shielded neural
policy, sometimes by a factor of two or more (e.g., {\sf Cartpole}, 
{\sf Self-Driving}, and {\sf Oscillator}). Our result demonstrates that 
executing a neural policy in tandem with a program distilled 
from it can retain performance, provided by the neural policy, while 
maintaining safety, provided by the verified program.}

{Although our synthesis algorithm does not guarantee convergence to 
the global minimum when applied to nonconvex RL problems, the 
results given in Table~\ref{tbl:result1} indicate that 
our algorithm can often produce high-quality control programs, with 
respect to a provided sketch, that converge reasonably fast.
In some of our benchmarks, however, the number of interventions are
significantly higher than the number of neural controller failures,
\emph{e.g.,} {\sf 8-Car platoon} and {\sf Oscillator}.  However, the high 
number of interventions is not primarily because of non-optimality of the 
synthesized programmatic controller. 
Instead, inherent safety issues in the neural network models are the main
culprit that triggers shield interventions. 
In {\sf 8-Car platoon}, after corrections made by our deterministic program, 
the neural model again takes an unsafe action in the next execution step 
so that the deterministic program has to make another correction. It is only 
after applying a number of such shield interventions that the system navigates 
into a part of the state space that can be safely operated on by the neural model.
For this benchmark, all system states where there occur shield interventions are 
indeed unsafe. We also examined the unsafe simulation runs made by executing 
the neural controller alone in {\sf Oscillator}. Among the large number of shield 
interventions (as reflected in Table~\ref{tbl:result1}), 74\% of them are 
effective and indeed prevent an unsafe neural decision.
Ineffective interventions in {\sf Oscillator} are due to the fact that, 
when optimizing equation~\eqref{eq:randomsearch}, a large penalty is given to 
unsafe states, causing the synthesized programmatic policy to weigh safety 
more than proximity when there exist a large number of unsafe neural decisions.
}

{\paragraph{Suitable Sketches.} 

Providing a suitable sketch may need domain knowledge. 
To help the user more easily tune the shape of a sketch, our approach 
provides algorithmic support by not requiring conditional statements 
in a program sketch and syntactic sugar, \emph{i.e.,} the user can simply 
provide an upper bound on the degree of an invariant sketch.

\begin{footnotesize}
\begin{table}[t]
\caption{Experimental Results on Tuning Invariant Degrees. TO means that
an adequate inductive invariant cannot be found within 2 hours.}
\label{tbl:result3}
\begin{tabular}{| c c c c c|}
    \hline
\rowcolor{black!20} Benchmarks & Degree & Verification & Interventions & Overhead \\
\multirow{3}{*}{Pendulum} & 2 & TO & - & - \\
& 4 & 22.6s & 40542 & 7.82\% \\
& 8 & 23.6s & 30787 & 8.79\% \\
\hline
\multirow{3}{*}{Self-Driving} & 2 & TO  & - & - \\
& 4 & 24s  & 128851 & 6.97\% \\
& 8 & 25.1s  & 123671 & 26.85\% \\
\hline
\multirow{3}{*}{8-Car-platoon} & 2 & 172.9s  & 43952 & 8.36\% \\
& 4 & 540.2s  & 37990 & 9.19\% \\
& 8 & TO  & - & - \\
\hline
\end{tabular}
\end{table}
\end{footnotesize}

Our experimental results are collected using the invariant sketch
defined in equation~\eqref{eq:invariantsketch} and we chose
an upper bound of 4 on the degree of all monomials included
in the sketch. Recall that invariants may also be used as
conditional predicates as part of a synthesized program.
We adjust the invariant degree upper bound to evaluate its effect 
in our synthesis procedure. The results are given in Table~\ref{tbl:result3}. 

\begin{small}
\begin{table*}[t]
\caption{Experimental Results on Handling Environment Changes.}
\label{tbl:result2}
\begin{tabular}{|l | c | c  c | c c c  c|}
    \hline
   \rowcolor{black!20} &  &  \multicolumn{2}{c|}{Neural Network} & \multicolumn{4}{c|}{Deterministic Program as Shield}  \\ 
   \rowcolor{black!20} \multirow{-2}{*}{Benchmarks} & \multirow{-2}{*}{Environment Change} &  Size & Failures & Size & Synthesis & Overhead & Shield Interventions \\ \hline
Cartpole & Increased Pole length by 0.15m & $1200\times900$ & 3 & 1  & 239s & 2.91\% & 8 \\
Pendulum & Increased Pendulum mass by 0.3kg & $1200\times900$ & 77 & 1  & 581s & 8.11\% & 8748 \\
Pendulum & Increased Pendulum length by 0.15m & $1200\times900$ & 76 & 1  & 483s & 6.53\% & 7060 \\
Self-driving & Added an obstacle that must be avoided  & $1200\times900$ & 203 & 1  & 392s & 8.42\% & 108320 \\
\hline
\end{tabular}
\end{table*}
\end{small}

Generally, high-degree invariants lead to fewer interventions because
they tend to be more permissive than low-degree ones.
However, high-degree invariants 
take more time to synthesize and verify. This is particularly true for high-dimension 
models such as {\sf 8-Car platoon}. Moreover, although high-degree invariants tend to 
have fewer interventions, they have larger overhead. For example, using a shield
of degree 8 in the {\sf Self-Driving} benchmark caused an overhead of 26.85\%.
This is because high-degree polynomial computations are time-consuming. 
On the other hand, an insufficient degree upper bound may not be permissive 
enough to obtain a valid invariant. It is, therefore, essential to consider the 
tradeoff between overhead and permissiveness when choosing the degree 
of an invariant.}

\paragraph{Handling Environment Changes.}
We consider the effectiveness of our tool when previously trained
neural network controllers are deployed in environment contexts
different from the environment used for training.  Here we consider
neural network controllers of larger size (two hidden layers with
$1200\times900$ neurons) than in the above experiments. This is
because we want to ensure that a neural policy is trained to be near
optimal in the environment context used for training.  These larger
networks were in general more difficult to train, requiring at least
1500 seconds to converge.

Our results are summarized in Table~\ref{tbl:result2}.  
{When the underlying environment sightly changes, learning
a new safety shield takes substantially shorter time than
training a new network.}
For {\sf Cartpole}, we simulated the trained controller in a new environment
  by increasing the length of the pole by 0.15 meters. The
neural controller failed 3 times in our 1000 episode
simulation; the shield interfered with the network operation
only 8 times to prevent these unsafe behaviors. The new shield was
synthesized in 239s significantly faster than retraining a new
neural network for the new environment.  For {\sf (Inverted)
  Pendulum}, we deployed the trained neural network in an environment
in which the pendulum's mass is increased by 0.3kg.  The neural
controller exhibits noticeably higher failure rates than in the
previous experiment; we were able to synthesize a safety shield
adapted to this new environment in 581 seconds that prevented these
violations.  The shield intervened with the operation of the network
only 8.7 number of times per episode. 
Similar results were observed when we increased the pendulum's length
by 0.15m.
For {\sf Self-driving}, we additionally required the car to avoid an obstacle. 
The synthesized shield provided safe actions to ensure collision-free motion.


\section{Related Work}
\label{sec:related}

\noindent{\bf Verification of Neural Networks.}  While neural networks
have historically found only limited application in safety- and
security-related contexts, recent advances have defined suitable
verification methodologies that are capable of providing stronger
assurance guarantees.  For example, Reluplex~\cite{reluplex} is an SMT
solver that supports linear real arithmetic with ReLU constraints and
has been used to verify safety properties in a collision avoidance
system.  AI$^{2}$~\cite{AI2,AI3} is an abstract interpretation tool
that can reason about robustness specifications for deep convolutional
networks. Robustness verification is also considered
in~\cite{nori,xiaowei}.  Systematic testing techniques such
as~\cite{DeepTest,DeepConcTest,featureguidedtesting} are designed to
automatically generate test cases to increase a coverage metric,
\emph{i.e.}, explore different parts of neural network architecture by
generating test inputs that maximize the number of activated neurons. 
These approaches ignore effects induced by the actual
environment in which the network is deployed.
{They are typically ineffective for safety verification of a neural network
controller over an infinite time horizon with complex system dynamics.
}
Unlike these whitebox efforts that reason over the
architecture of the network, our verification framework is fully
blackbox, using a network only as an oracle to learn a simpler
deterministic program. This gives us the ability to reason about
safety entirely in terms of the synthesized program, using a shielding
mechanism derived from the program to ensure that the neural policy
can only explore safe regions.


\noindent{\bf Safe Reinforcement Learning.}  The machine learning
community has explored various techniques to develop safe
reinforcement machine learning algorithms in contexts similar to ours,
\emph{e.g.},~\cite{safemdp,safemdpgp}.  In some cases, verification is
used as a safety validation oracle~\cite{reachabilitygp,safemodelrl}.
In contrast to these efforts, our inductive synthesis algorithm can be
freely incorporated into any existing reinforcement learning framework
and applied to any legacy machine learning model.  More importantly,
our design allows us to synthesize new shields from existing ones,
without requiring retraining of the network, under reasonable changes
to the environment.

\noindent{\bf Syntax-Guided Synthesis for Machine Learning.}  An
important reason underlying the success of program synthesis
is that sketches of the desired program~\cite{sketch,sketchthesis},
often provided along with example data, can be used to effectively
restrict a search space and allow users to provide additional insight
about the desired output~\cite{sygus}.  Our sketch based inductive
synthesis approach is a realization of this idea applicable to
continuous and infinite state control systems central to many machine
learning tasks.  A high-level policy language grammar is used in our
approach to constrain the shape of possible synthesized deterministic
policy programs.
{Our program parameters in sketches are continuous, because they 
naturally fit continuous control problems. However, our synthesis 
algorithm (line 9 of Algorithm~\ref{learnverifiedcontroller}) can call
a derivate-free random search method (which does not require the 
gradient or its approximative finite difference) for synthesizing functions 
whose domain is disconnected, (mixed-) integer, or non-smooth.
Our synthesizer thus may be applied to other domains that are not
continuous or differentiable like most modern program synthesis 
algorithms~\cite{sketch,sketchthesis}.}

In a similar vein, recent efforts on interpretable machine learning
generate interpretable models such as program code~\cite{pirl}
or decision trees~\cite{qdagger} as output, after which traditional
symbolic verification techniques can be leveraged to prove
program properties.  
Our approach novelly ensures that only safe programs are synthesized
via a CEGIS procedure and can provide safety guarantees on the
original high-performing neural networks via invariant inference. 
A detailed comparison was provided on page~\pageref{interp-comp}.



{\noindent{\bf Controller Synthesis.}
Random search~\cite{randomsearchbasic} has long been utilized 
as an effective approach for controller synthesis. In~\cite{NIPS2018}, 
several extensions of random search have been proposed that 
substantially improve its performance when applied to reinforcement 
learning. However, \cite{NIPS2018} learns a single linear controller that 
we found to be insufficient to guarantee safety for our benchmarks in 
our experience. We often need to learn a program that involves a family 
of linear controllers conditioned on different input spaces to guarantee safety.
The novelty of our approach is that it can automatically synthesize programs 
with conditional statements by need, which is critical to ensure safety.

In Sec.~\ref{sec:eval}, linear sketches were used in our experiments 
for shield synthesis. LQR-Tree based approaches such as~\cite{RSS2009} 
can synthesize a controller consisting of a set of linear quadratic 
regulator (LQR~\cite{LQR}) controllers each applicable in a different 
portion of the state space. 
These approaches focus on stability while our approach primarily 
addresses safety. In our experiments, we observed that because LQR 
does not take safe/unsafe regions into consideration, synthesized 
LQR controllers can regularly violate safety constraints.}

\noindent{\bf Runtime Shielding.}  
Shield synthesis has been used in runtime enforcement for
reactive systems~\cite{shield1,shield2}.
Safety shielding in reinforcement learning was first
introduced in~\cite{AB+18}.
{
Because their verification approach is not symbolic, however,
it can only work over finite discrete state and action systems.  Since
states in infinite state and continuous action systems are not
enumerable, using these methods requires the end user to provide a
finite abstraction of  complex environment dynamics; such
abstractions are typically too coarse to be useful (because they
result in too much over-approximation), or otherwise have too many
states to be analyzable~\cite{realsyn}.  Indeed, automated environment
abstraction tools such as~\cite{tulip} often take hours even for
simple 4-dimensional systems.  Our approach embraces the nature of 
infinity in control systems by learning a \emph{symbolic} shield from an 
inductive invariant for a synthesized program that includes an infinite number 
of environment states under which the program can be guaranteed to be
safe.  We therefore believe our framework provides a more promising
verification pathway for machine learning in high-dimensional control systems.}
In general, these efforts focused on shielding
of discrete and finite systems with no obvious generalization to
effectively deal with continuous and infinite systems that our
approach can monitor and protect.
{Shielding in continuous control systems was introduced in~\cite{CDC2014,ICRA2012}.
These approaches are based on HJ reachability (Hamilton-Jacobi 
Reachability~\cite{HJReachability}). 
However, current formulations of HJ reachability are limited to systems with 
approximately five dimensions, making high-dimensional system verification 
intractable. Unlike their least restrictive control law framework that decouples 
safety and learning, our approach only considers state spaces that a learned 
controller attempts to explore and is therefore capable of synthesizing safety 
shields even in high dimensions.}


\section{Conclusions}
\label{sec:conc}

This paper presents an inductive synthesis-based toolchain
that can verify neural network control policies
within an environment formalized as an infinite-state transition system. 
The key idea is a novel synthesis
framework capable of synthesizing a deterministic policy program based
on an user-given sketch that resembles a neural policy oracle and
simultaneously satisfies a safety specification using a
counterexample-guided synthesis loop.  Verification soundness is
achieved at runtime by using the learned deterministic program along
with its learned inductive invariant as a shield to protect the neural
policy, correcting the neural policy's action only if the chosen
action can cause violation of the inductive invariant.  Experimental
results demonstrate that our approach can be used to realize fully
trustworthy reinforcement learning systems with low overhead.


\bibliography{draft}

\end{document}